\documentclass[conference]{IEEEtran}
\usepackage{cite}
\usepackage{amsmath,amssymb,amsfonts}
\usepackage{multicol,caption}
\usepackage{commath}
\usepackage{algorithmic}
\usepackage[ruled]{algorithm2e}
\usepackage{textcomp}
\usepackage{multirow}
\usepackage{xcolor}
\usepackage{cite}
\usepackage{amsmath,amssymb,amsfonts}
\usepackage{comment}
\usepackage{graphicx}
\usepackage{textcomp}
\usepackage{xcolor}
\usepackage[none]{hyphenat}
\usepackage{times}
\usepackage{epsfig}
\usepackage{graphicx}
\usepackage{balance}
\usepackage[utf8]{inputenc}
\usepackage{booktabs}
\usepackage{graphicx}% http://ctan.org/pkg/graphicx
\usepackage{array}% http://ctan.org/pkg/array
\DeclareMathOperator*{\argmax}{arg\,max}

\def\BibTeX{{\rm B\kern-.05em{\sc i\kern-.025em b}\kern-.08em
    T\kern-.1667em\lower.7ex\hbox{E}\kern-.125emX}}

\title{AdvART: Adversarial Art for Camouflaged Object Detection Attacks}
\author{\IEEEauthorblockN{ Amira Guesmi$^1$, Ioan Marius Bilasco$^2$, Muhammad Shafique$^1$, and Ihsen Alouani$^3$}%\\
 \IEEEauthorblockA{$^1$\textit{eBrain Lab, Division of Engineering, New York University (NYU) Abu Dhabi, UAE} \\
 \textit{$^2$CRIStAL, Univ. Lille, CNRS, Centrale Lille, UMR 9189, France}\\
 \textit{$^3$CSIT, Queen’s University Belfast, UK} \\
 } }
\begin{document}
\maketitle
\begin{abstract}
Physical adversarial attacks pose a significant practical threat as it deceives deep learning systems operating in the real world by producing prominent and maliciously designed physical perturbations. Emphasizing the evaluation of naturalness is crucial in such attacks, as humans can readily detect and eliminate unnatural manipulations. To overcome this limitation, recent work has proposed leveraging generative adversarial networks (GANs) to generate naturalistic patches, which may not catch human's attention. However, these approaches suffer from a limited latent space which leads to an inevitable trade-off between naturalness and attack efficiency. In this paper, we propose a novel approach to generate naturalistic and inconspicuous adversarial patches.  
Specifically, we redefine the optimization problem by introducing an additional loss term to the cost function. This term works as a semantic constraint to ensure that the generated camouflage pattern holds semantic meaning rather than arbitrary patterns. The additional term leverages similarity metrics to construct a similarity loss that we optimize within the global objective function. Our technique is based on directly manipulating the pixel values in the patch, which gives higher flexibility and larger space compared to the GAN-based techniques that are based on indirectly optimizing the patch by modifying the latent vector. Our attack achieves superior success rate 
of up to 91.19\% and 72\%, respectively, in the digital world and when deployed in smart cameras at the edge compared to the GAN-based technique.
\end{abstract}    
\section{Introduction}
\label{intro}

%DNN based real-time object detection systems are being increasingly deployed in safety-critical and security-sensitive domains such as video surveillance \cite{boudjit2022human,sabater2020robust,8524260}, self-driving cars \cite{al2017deep}, health-care \cite{miotto2018deep}, etc. However, DNNs suffer from a critical security vulnerability, researchers have demonstrated that these networks are vulnerable to adversarial perturbations \cite{CW, LiV15, adv_example, PapernotMJFCS15}. These malicious examples can even be physically produced in the real world \cite{EvtimovEFKLPRS17,phy9,phy10, neuroattack}. 

Physical attack scenarios \cite{phy9, phy10, guesmi2023saam, guesmi2023aparate} involve attackers designing printable adversarial patches for deployment in scenes captured by the victim model. These patches are not generated with constraints on noise magnitude, but rather on location and printability. Considering their practical application in real-world situations, physical patch-based attacks are more damaging and practical for real-life scenarios. This paper focuses on adversarial patches and specifically investigates their undetectability from a naturalistic distribution perspective.

\begin{figure}[!ht] %tp
\centering
\includegraphics[width=\columnwidth]{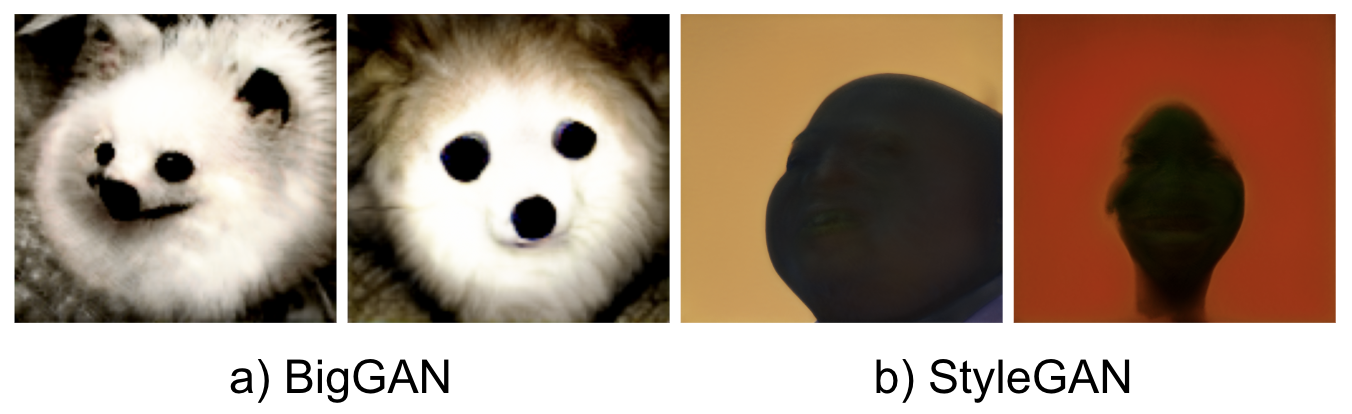}
\caption{Different results of different attempts to generate NAP patches \cite{Hu21} using different latent vector initialization for different GANs: a) BigGAN, and b) StyleGAN.}
\label{init}
\end{figure}

Prior research on adversarial patches for object detection \cite{guesmi2023physical, Eykholt2018, thys2019, Zhao2019} has primarily focused on enhancing attack performance and increasing the strength of adversarial noise. However, this approach often leads to the creation of conspicuous and noticeable patches that can be easily identified by human observers. To tackle this issue, some researchers have proposed the utilization of generative adversarial networks (GANs) \cite{Pavlitskaya22, Bai22, Hu21, Liu20} to generate naturalistic adversarial patches. However, these methods were proven to suffer from an extremely limited latent space \cite{Hu21, dap}, which limits the performance of the generated patch especially when trying to incorporate multiple transformations in the optimization process. In addition, the generation of a naturalistic pattern is not always guaranteed. As shown in Figure \ref{init}, with different initialization the GAN-based framework converges to unrealistic patterns.

To overcome this limitation and to generate effective attacks with naturalistic patterns, we propose a novel framework (AdvART). %(presented in Figure \ref{approach}). 
Our approach involves optimizing towards a target natural image while simultaneously maximizing the loss of the victim model, aimed to undermine detection systems. This approach involves an optimization process utilizing a novel objective function that incorporates a semantic constraint to guide the pattern of the generated noise and converge to a semantically meaningful adversarial pattern. %Figure \ref{advartvssota} showcases illustrations of the resulting adversarial patches printed on a T-shirt, demonstrating their effectiveness.

\noindent\textbf{Novel contributions -- } The main contributions of this paper are summarized as follows:
\begin{itemize}
    %\item We show that GAN-based approaches do not always converge to realistic patterns (Section \ref{gan-based}).
    \item We propose a novel framework (AdvART) that generates naturalistic patches while retaining attack performance. Our technique incorporates a semantic constraint to force the optimized noise to follow a predefined natural/artistic pattern. %while giving higher flexibility (compared to GAN-based techniques) to incorporate multiple transformations aimed at increasing the robustness of the generated patch. %We illustrate our approach with AdvART, where the adversarial patch looks like artistic paintings.
    %We introduce a groundbreaking framework, AdvART, which excels in producing naturalistic patches while preserving strong attack performance. Our approach seamlessly integrates a semantic constraint, compelling the optimized noise to conform to a predefined artistic pattern. 
    \item Our proposed technique offers greater flexibility compared to GAN-based techniques, allowing for the incorporation of multiple transformations. These transformations are strategically aimed at enhancing the overall robustness of the generated patch.
    
    \item We conduct a comprehensive analysis of the effectiveness of the proposed approach by evaluating its performance and attack success rate in terms of mean average precision (mAP), as well as examining the patch's transferability across different detectors. %Our patch achieves an attack success rate of 91.19\%, 65.07\%, 82.3\%, and 84.18\% in the digital world (INRIA dataset) when targeting the YOLOv4tiny, YOLOv3tiny, YOLOv4, and YOLOv3 detectors, respectively for edge systems. 
    Our patch achieves a mAP of 8.81\%, 34.93\%, 17.7\%, and 15.82\% in the digital world (INRIA dataset \cite{inria}) when targeting the YOLOv4tiny, YOLOv3tiny, YOLOv4 \cite{yolov4}, and YOLOv3 \cite{yolov3} detectors, respectively for edge systems. 
    \item We demonstrate the effectiveness of our proposed patch in real-world scenarios, achieving a success rate of 72\%. % showcasing its effectiveness in the digital world, as well as when printed on paper and t-shirts.
\end{itemize}

\section{ Related Work}
\label{sec:related}
\begin{figure*} %[!t]
\centering
\includegraphics[width=2\columnwidth]{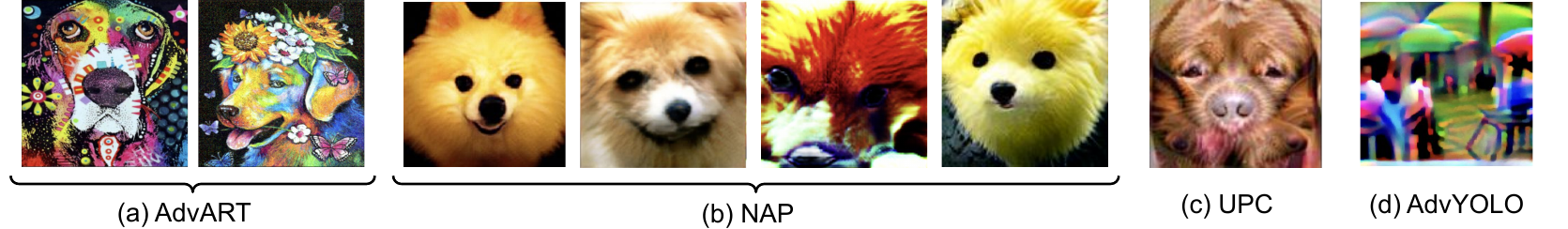}
\caption{AdvART patch vs State-of-the-Art patches: (a) AdvART patch, (b) NAP \cite{Hu21}, (c) UPC patch \cite{Huang2020}, and (d) AdvYOLO \cite{thys2019}.}
\label{advartvssota}
\end{figure*}
At first, the proposed physical attacks aiming at fooling person detectors were generated without any constraints on the patch appearance \cite{thys2019 , Lee2019 , Xu2020 , Wu2020}. Their main focus was on performance and producing effective attacks without any focus on attack stealthiness.
To overcome this limitation, some works proposed to leverage the learned image manifold of pre-trained Generative Adversarial Networks (GANs) upon real-world images.  Authors in \cite{Hu21} proposed a naturalistic patch (NAP) based on two GANs (i.e., BigGAN \cite{biggan} and StyleGAN \cite{karras2020analyzing}). The training on GANs needs a lot of resources in terms of computing units and time to converge if they converge to realistic images in the first place.  Authors in \cite{Huang2020} also proposed a universal camouflage pattern (UPC) that is visually similar to natural images.
In \cite{Pavlitskaya22}, authors tried to extend the existing work to make the patch suppress objects near the patch. However, what they proposed is unrealistic in terms of the positioning of the patch and its size.
Also, another limitation of these approaches is that the patch will be restricted to objects that the GAN was trained to produce, these approaches may not always converge to a natural looking patch. 
A comparison of AdvART with state-of-the-art attacks is provided in Table \ref{Tab:related}. In Figure \ref{advartvssota}, we illustrate different state of the art adversarial patches in addition to ours (AdvART-based).

\begin{table}[!ht]
%\scalebox{0.8}{
  \centering
\caption{AdvART vs State-of-the-Art adversarial patches. (EOT: Expectation Over Transformations, TV: Total Variation, NPS: Non-Printability Score).}\label{Tab:related}
  \begin{tabular}{lccc}
    \hline
    \textbf{Attack}             & \textbf{Robustness } & \textbf{Stealthiness} &\textbf{Form}\\
    \hline
      \textbf{AdvYOLO \cite{thys2019} } & EOT, TV, NPS  &  (-)               & Cardboard \\
     \textbf{UPC\cite{Huang2020} }               &  EOT, TV      &  $L_\infty$ norm & Clothing \\
      \textbf{NAP \cite{Hu21}}                   & TV            & GAN           &  Clothing\\
      \textbf{AdvART (ours)}                     & EOT, TV   & $L_{sim}$  &  Clothing \\
  \hline
\end{tabular}
\end{table}
%}

\section{AdvART: Our Adversarial Art Framework for Object Detection Attacks } %Proposed Approach
\label{sec:proposed}
%============================
\subsection{Problem formulation}
%============================

In an object detection context, given a benign image $I$, the purpose of the adversarial attack is to jeopardize the object detector and suppress the target objects using the maliciously designed adversarial example $I^*$.
Technically, the adversarial example with a generated patch can be formulated as:

\begin{equation}
    I^* = (1 - M_P) \odot I + M_P \odot P
\end{equation}
$\odot$ is the component-wise multiplication, $P$ is the adversarial patch, and $M_P$ is the patch mask, used to constrict the size, shape, and location of the adversarial patch. 
The problem of generating an adversarial example can be formulated as a constrained optimization \ref{eq:adv}, given an original input image $I$ and an object detector $ F(.) $:
\begin{equation}
\label{eq:adv}
    \min_{P} \left\|P\right\|_p  \\
    ~ s.t. F((1 - M_P) \odot I + M_P \odot P) \neq F(I)\\
    %d_{adv} \neq d
\end{equation}

The objective is to find a minimal adversarial noise, $P$, such that when placed on an arbitrary object from a target input domain $U$, it will selectively jeopardize the underlying DNN-based model $F(.)$ by making objects undetectable. 
Note that one cannot find a closed-form solution for this optimization problem since the DNN-based model $F(.)$ is a non-convex machine learning model. Therefore, Equation \ref{eq:adv} can be formulated as follows to numerically solve the problem using empirical approximation techniques:

\begin{equation}
\label{eq:formulation}
    \argmax_{P} \sum_{I \in \mathcal{U}} l(F((1 - M_P) \odot I + M_P \odot P), F(I)
\end{equation}
where $l$ is a predefined loss function. %and $\mathcal{U} \subset U$ is the attacker’s training dataset.
We can use existing optimization techniques (e.g., Adam \cite{adam}) to solve this problem. In each iteration of the training, the optimizer updates the adversarial patch $P$.

%============================
\subsection{Overview of Our Framework}
%============================
\begin{figure*}[!ht]
\centering
\includegraphics[width=1.4\columnwidth]{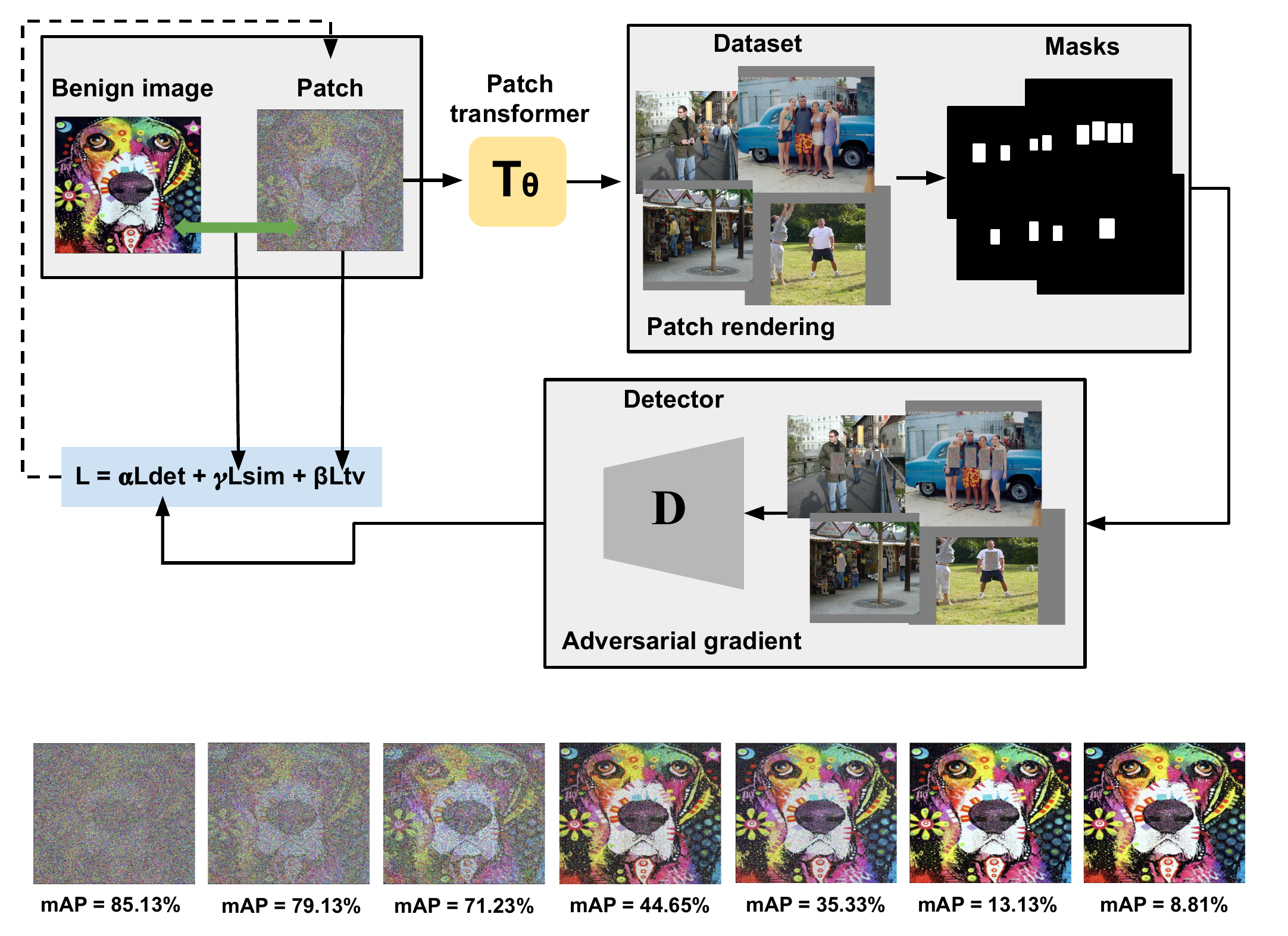}
\caption{Upper: Overview of the proposed framework: % We optimize the adversarial patch $P$ in a way that when placed on a target object, this latter becomes undetectable by an object detector while maintaining the naturalistic and artistic pattern which will be guided by a given benign image.We first feed the patch to the patch transformer to perform the \textit{geometric transformations} to make the patch more robust. We, later on, render the patch on top of the input image using the generated masks. After that, we feed the resulting adversarial image to the object detector to compute different loss functions. Then, we compute the gradient of the patch and based on this information we update the patch $P$. 
Bottom: Evolution of patch's appearance with the corresponding achieved mean average precision.}
\label{approach}
\end{figure*}

The primary objective of our research is to generate physical adversarial patches that effectively deceive DNN-based object detectors, without being conspicuous or attention-grabbing while being robust to real-world constraints. In this study, we place our focus on simultaneously fulfilling three essential requirements crucial for real-world applicability. These requirements are as follows:
\begin{itemize}
 \item \textbf{Patch effectiveness}: Ensuring the generated patch significantly degrades the performance of the detector. 
 \item \textbf{Patch stealthiness}: Ensuring that the patch remains inconspicuous to human observers and does not attract attention.  
 \item \textbf{Patch robustness}: Maintaining the patch's attack capability in dynamic environments, including resilience to physical constraints.
\end{itemize}
To achieve these goals, we propose an objective function that incorporates a semantic constraint (i.e., similarity loss), which guarantees the artistic pattern of the adversarial patch. Additionally, a detection loss is employed to maintain the attack's performance while considering complex physical factors such as object size, shape, and location.

Figure \ref{approach} provides an overview of the proposed framework. Our aim is to optimize the adversarial patch $P$ in such a way that the target object becomes undetectable by an object detector, while ensuring the patch retains a naturalistic and artistic pattern guided by a given benign image.

To achieve this, we employ several steps. Firstly, we pass the patch through a patch transformer, applying geometric transformations to enhance its robustness. Next, we overlay the patch onto the input image using generated masks. The resulting adversarial image is then fed into the object detector to compute the detection loss, similarity loss, and total variation.

Using the gradient information obtained from the patch, we update the patch $P$. \textit{Our goal is to generate physical adversarial patches that possess a natural appearance while maintaining their effectiveness in evading detection.} This is achieved through iterative gradient updates performed on the adversarial patch $(P)$ in the pixel space, optimizing our objective defined as follows:

\begin{equation}
    Loss_{total} = \alpha L_{det} + \beta L_{sim}^2 + \gamma L_{tv}
\end{equation}

Where $L_{det}$ is the adversarial detection loss (See Section \ref{Ldet}). $L_{sim}$ is the similarity loss (See Section \ref{stealth}).
$L_{tv}$ is the total variation loss on the generated image to encourage smoothness (See Section \ref{robustness}).
$\alpha$, $\beta$, and $\gamma$ are hyper-parameters used to scale the three losses. For our experiments we set $\alpha = 1$, $\beta = 8$, and $\gamma = 0.5$
We optimize the total loss using Adam \cite{adam} optimizer. We try to minimize the object function $L_{total}$ and optimize the adversarial patch. We freeze all weights and biases in the object detector, and only update the pixel values of the adversarial patch. %The patch is randomly initialized. 

%============================
\subsection{Patch effectiveness}
%============================
\label{Ldet}
Object detectors like YOLO \cite{yolo, yolov2, yolov3, yolov4}produce multiple boxes or detections. Our objective is to target two specific quantities for each detection $j$: the objectness probability $D^j_{obj}$ and the class probability $D^j_{cls}$. By minimizing the objectness probability $D^j_{obj}$, we aim to prevent the j-th object from being detected altogether. Conversely, minimizing the class probability $D^j_{cls}$ results in the j-th object being misclassified into a different class, such as a person being classified as a dog. In this paper, our focus is on targeting the person class. Therefore, we minimize both the objectness $D^j_{obj}$ and class probabilities $D^j_{cls}$ specifically for the person class. To speed up the iterations, we do not compute the loss over all detected boxes. Instead, we only consider the detected box with the highest objectness and class probabilities. Our adversarial detection loss is defined below:
\vspace{-3mm}
\begin{equation}
    L_{det} = \frac{1}{N} \sum^{N}_{i = 1} max_j [D^j_{obj} (I'_i)D^j_{cls}(I'_i)]
\end{equation}

%============================
\subsection{Patch Stealthiness}
%============================
\label{stealth}
%\subsection{Adversarial Similarity Loss}
%project into the onto the surface of $L_p$ norm-balls with radius $\epsilon$ and centered at I
%We constraint the pattern of the generate patch to look similar to a predefined artistic pattern. in order to do that we use a semantic constraint where we choose N as natural artistic image to ensure the generated camouflage pattern is semantically meaningful. 
%The key idea to generate a patch similar to existing piece of art, and this accomplished by create a new loss term that represents the distance of the patch to the art piece. The proposed similarity loss is designed to minimize the distance between a benign artistic image and the patch.

%where we choose N as natural artistic image to ensure the generated camouflage pattern is semantically meaningful.
%%
To ensure that the generated patch exhibits a pattern similar to a predefined natural or artistic pattern, we impose a constraint on its appearance. This is achieved through the use of a semantic constraint, where we select a natural/ artistic image denoted as N, and we force the adversarial noise to follow the artistic pattern rather than having meaningless patterns and ensure that the generated camouflage pattern holds semantic meaning. The primary concept behind achieving a patch similar to an existing piece of art involves introducing a new loss term. This loss term quantifies the distance between the patch and the art piece, aiming to minimize this distance. Referred to as the similarity loss, it is specifically designed to reduce the disparity between the patch and a benign artistic image. By integrating this loss term, we facilitate the generation of a patch that closely resembles the artistic pattern. As depicted in Figure \ref{approach}, which illustrates the changes in patch appearance during training and its corresponding impact on object detector performance in terms of mean average precision, naturalness is improved in conjunction with attack efficiency.

$L_{sim}$ is the similarity loss between the target natural image $N$ and the adversarial patch $P$. It is defined as:\\
\textbf{MSE-based similarity loss}
\begin{equation}
    L_{sim} = \frac{1}{n} \sum_{i,j} ( P_{i,j} - N_{i,j} )^2 
\end{equation}
%where $m$ and $n$ are the dimensions 
\textbf{Cosine similarity-based similarity loss}

\begin{equation}
    L_{sim} = -\left(\frac{\sum_{i,j}P_{i,j} N_{i,j}}{\sqrt{\sum_{i,j}P^2_{i,j}}\sqrt{\sum_{i,j}N^2_{i,j}}} \right)
\end{equation}

We square the similarity loss term, to slow the rate of increase (the slope or the rate of change) and delay the convergence of the similarity metric with respect to the detection loss. %The similarity metric is a linear function compared to the detection metric which is a non-linear function that converges much slower. 

%\begin{figure*}[!ht]
%\centering
%\includegraphics[width=2\columnwidth]{figures/evolution.pdf}
%\caption{Changes in patch appearance during training and its corresponding impact on object detector performance in terms of mean average precision.}
%\label{evolution}
%\end{figure*}

%To find a trade-off between naturalness/stealthiness, efficiency and robustness of the generated adversarial patches..
%============================
\subsection{Patch Robustness}
%============================
\label{robustness}
%============================
\subsubsection{Expectation Over Transformation (EOT) \cite{EOT}}
%============================
We incorporate the physical world variables throughout the adversarial patch optimization process to increase its robustness and retain the same performance in realistic scenarios. Various factors, such as fluctuating lighting, various viewpoints, noise, etc., are frequently present in real-world scenarios. We use several physical transformations to simulate these dynamic factors. Technically, we take into account the transformations of the inclusion of variable conditions, such as adding noise, random rotation, variable scales, lighting variation, etc. The above physical transformation operations are performed using the patch transformer. The geometric transformations applied are: \textbf{Random scaling} of the patch to a size that is roughly equivalent to its physical size in the scene. Perform \textbf{random rotations} ($\pm20^\circ$) on the patch $P$ about the center of the bounding boxes $B_{i,k}^U$. The above simulate placement and printing size uncertainties. The color space transformations are done by changing the pixel intensity values by adding \textbf{random noise} ($\pm 0.1$ ), performing \textbf{random contrast} adjustment of the value ($[\ 0.8, 1.2 ]\ $), and \textbf{random brightness} adjustment ($\pm 0.1$). %Let $\Tilde{I}_i$ be $I_i$ embedded with P following the steps above.
The resulting image $T_\theta (I_i)$ is forward propagated through the Object Detector.

%============================
\subsubsection{Total Variation Norm (TV loss)}
%============================
The characteristics of natural images include smooth and consistent patches with gradual color changes within each patch \cite{mahendran2015understanding}. Therefore, to increase the plausibility of physical attacks, smooth and consistent perturbations are preferred. %Additionally, extreme differences between adjacent pixels in the perturbation may not be accurately captured by cameras due to sampling noise. This means that non-smooth perturbations may not be physically realizable \cite{Sharif2016FaceRecognitionAttacks}. 
To address this issue, the total variation (TV) \cite{mahendran2015understanding} loss is introduced to maintain the smoothness of the perturbation.
For a perturbation $P$, TV loss is defined as:
%$L_{tv} = \sum_{i,j} \sqrt{(P_{i+1,j} - P_{i,j})^2 + (P_{i,j+1} - P_{i,j})^2}$
\begin{equation}
    L_{tv} = \sum_{i,j} \sqrt{(P_{i+1,j} - P_{i,j})^2 + (P_{i,j+1} - P_{i,j})^2}
\end{equation}

Where the subindices $i$ and $j$ refer to the pixel coordinate of the patch $P$.

%\input{sec/methodology}
%============================
\section{Evaluation of attack performance}
%============================
\label{sec:eval}
\begin{table}[!ht]
  \centering
%\small
  \caption{Transferability analysis in terms of mAP of AdvART on INRIA dataset using different detectors.}
  \label{performance}
  \begin{tabular}{lccccc}
    \hline
    \textbf{Models} &\textbf{Yolov3}  & \textbf{Yolov3tiny} & \textbf{Yolov4}  & \textbf{Yolov4tiny} \\
    \hline
      %\textbf{YOLOv2}        & 35.67$\%$  & 37.61$\%$  & 36.20$\%$  & 65.43$\%$ \\
      \textbf{ Yolov3 \cite{yolov3}}       &\textbf{15.82$\%$}   & 52.77$\%$  &29.37$\%$  &58.24$\%$ \\
      \textbf{Yolov3tiny}  &  41.46$\%$ & \textbf{34.93$\%$}  &  37.46$\%$ & 56.29$\%$ \\
      \textbf{Yolov4 \cite{yolov4}}      & 27.18 $\%$ &  41.36$\%$ & \textbf{17.7$\%$}  & 65.84$\%$\\
      \textbf{Yolov4tiny}  & 60.70$\%$  & 57.40$\%$  & 47.46 $\%$ & \textbf{8.81$\%$}\\
  \hline
\end{tabular}
\end{table}

We use the mean average precision (mAP) as our evaluation metric. Same as in \cite{thys2019, Hu21}, we use each detector's detection boxes on the clean dataset as the ground truth boxes (i.e., if there is no adversarial patch, the mAP of detectors will be $100\%$) and we report the mean average precision (mAP) when adding the adversarial patches. Table \ref{performance} shows the evaluation results on the INRIA dataset. We use four different detectors to generate the patches. The horizontal detectors are the victim ones and the verticals are the ones used for patch generation.

As shown in Table \ref{performance}, expectedly, when the victim detector used for training is the same one used for testing, we accomplish a great attack performance. %Additionally, our technique can also do well against detectors that were not used for training which proves the transferability across models of our attack. 
Importantly, Table \ref{performance} shows that our patch is significantly transferable to other models.
\begin{table}[!ht]
  \centering
%\small
  \caption{Attack performance in terms of mAP of AdvART vs state-of-the-art adversarial patches.}
  \label{comparison}
  \begin{tabular}{lcccc}
    \hline
    \textbf{Models} & \textbf{AdvART}& \textbf{NAP \cite{Hu21}} & \textbf{AdvYOLO \cite{thys2019}}& \textbf{UPC \cite{Huang2020}}\\
    \hline
      \textbf{Yolov2}      &  20.8$\%$  &   12.06$\%$ &  \textbf{2.13$\%$} & 48.62$\%$\\
      \textbf{Yolov3}      &  \textbf{15.82$\%$}  &   34.93 $\%$&  22.51$\%$  &   54.40$\%$\\
      \textbf{Yolov3tiny}  &  34.93$\%$  &   10.02$\%$&  \textbf{8.74$\%$}&     63.82$\%$\\
      \textbf{Yolov4}      &  17.7 $\%$ &   22.63$\%$&  \textbf{12.89$\%$}&   64.21 $\%$\\
      \textbf{Yolov4tiny}  &   8.81$\%$ &   8.67$\%$&   \textbf{3.25$\%$}&   57.93 $\%$\\
  \hline
\end{tabular}
\end{table}

%When comparing different techniques we notice a higher performance of other techniques compared to ours, it is worth noting that the results reported for NAP are for the case where the patch was trained without transformations and tested without performing any transformations, when training with transformations we notice a drop in patch performance (discussed in section A), same for the other techniques. However, the results reported for our patch are those when the patch is trained for different transformations. 

As shown in Table \ref{comparison}, when we compare our technique to others, we can notice that some previously proposed methods exhibit superior performance for some of the detectors. It's important to note that the reported results for the AdvYOLO \cite{thys2019} attack there were no constraint on the appearance of the patch which gave higher flexibility to attain higher attack performance.
Additionally, the NAP \cite{Hu21} technique pertain to a specific scenario where the patch was trained without any transformations and tested without applying any transformations. However, it's crucial to emphasize that, as discussed in Section \ref{sec:transformations}, when we train their patch with transformations with the aim of improving robustness, there is a noticeable decrease in patch performance. This phenomenon is observed in other techniques as well. Conversely, the results we report for our patch reflect a scenario where the patch is trained to accommodate various transformations. In this context, our patch undergoes training with the explicit inclusion of transformations, which is an essential factor contributing to its performance characteristics.
%=============================
\subsection{Cross-Dataset Evaluation}
%=============================
We use Yolov4tiny to train an adversarial patch on the INRIA dataset and test the produced patch on the MPII dataset. As shown in Table \ref{performance-cross}, our patch is transferable cross-datasets. An adversarial patch generated on the INRIA dataset is still effective on MPII \cite{MPII} with an mAP equal to $15.53\%$.
\begin{table}[!ht]
  \centering
%\small
  \caption{Attack performance (mAP) of AdvART on MPII.}
  \label{performance-cross}
  \begin{tabular}{lcc}
    \hline
    \textbf{Datasets} & \textbf{INRIA} & \textbf{MPII}\\
    \hline
      \textbf{mAP}  & 8.81$\%$  & 15.53 $\%$     \\
  \hline
\end{tabular}
\end{table}
\vspace{-3mm}
%============================
\section{Subjective Evaluation for the Naturalness of Different Adversarial Patches}
%============================
\begin{figure*}[!ht]
\centering
\includegraphics[width=1.6\columnwidth]{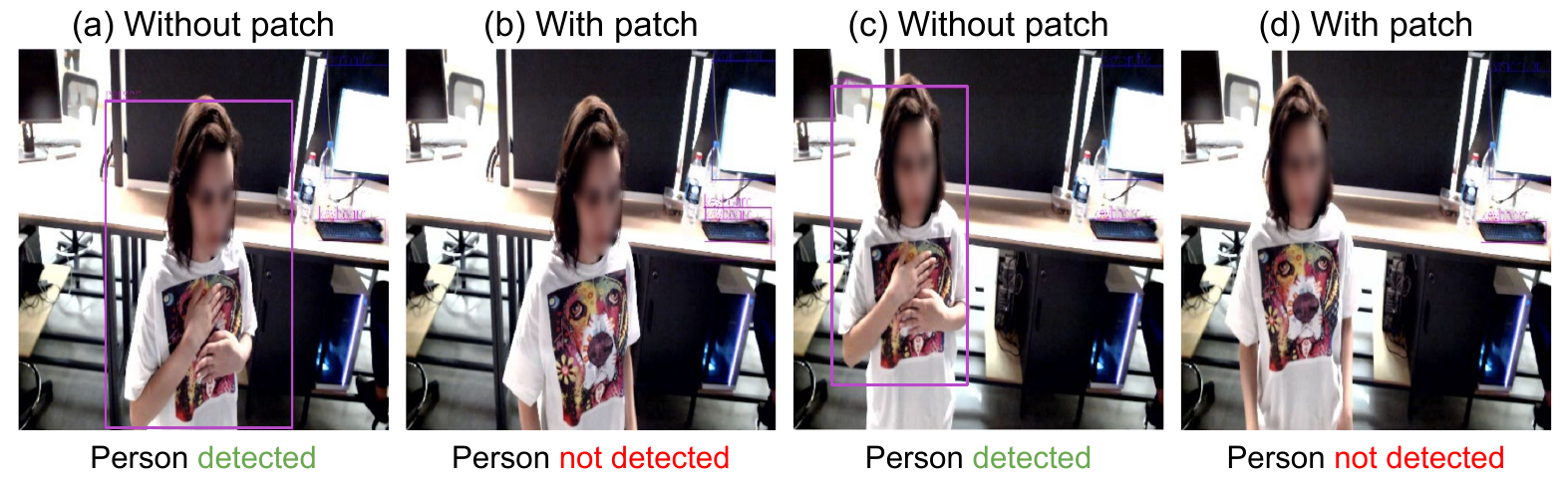}
\caption{Illustrations of AdvART patch performance when printed on a t-shirt for different view angles.}
\label{advarttshirt}
\end{figure*}
%The primary focus of our proposed approach lies in the naturalness and conspicuousness of the generated adversarial patch to human observers. To assess this aspect, we conducted a formal set of subjective evaluations, comparing the naturalness of our proposed patches with both baseline and real images. The subjective survey was administered, involving a group of 20 participants, to gather comprehensive feedback.

%The evaluation involved a diverse group of participants, including both males and females with different backgrounds, and they fall within the age range of 19 to 56 while having no prior knowledge about the topic. This approach helps minimize potential biases and enhances the credibility of our subjective evaluation. The participants were asked to provide a score between 0 and 100\% for both naturalness and the absence of conspicuous patterns in the provided image. As illustrated in Table \ref{tbl:nature}, our patch received the highest score corresponding to the highest naturalness and indicating that our patch is highly effective in blending seamlessly with the surroundings or context in which it was evaluated.

\begin{table}[htp!]
  \centering
    \caption{Subjective test for the naturalness evaluation of our adversarial patch with other baselines. The Naturalness scores are the average of percentages for each patch image over the whole group of participants. %As shown in the results, ours get more votes than others, demonstrating its effectiveness.
    }\label{tbl:nature}
  \begin{tabular}{ | c | c | c | c | c | }
    \hline
      Patches & AdvART & NAP \cite{Hu21}&  UPC \cite{Huang2020} & AdvYOLO \cite{thys2019}
        \\ \hline %%%%%%%%%%%%%%%%%%%%%%%%%%%%%%%%%%%%%%%%%%%%%%%%%%%%%%%%%%%%%%
    Images &
    \begin{minipage}{.08\textwidth}
      \includegraphics[width=15mm, height=15mm]{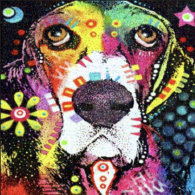}
    \end{minipage}
    &
    \begin{minipage}{.07\textwidth}
      \includegraphics[width=15mm, height=15mm]{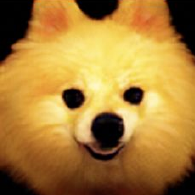}
    \end{minipage}
    & 
    \begin{minipage}{.07\textwidth}
      \includegraphics[width=15mm, height=15mm]{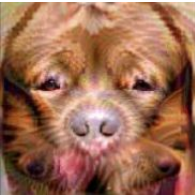}
    \end{minipage}
    & 
    \begin{minipage}{.1\textwidth}
      \includegraphics[width=15mm, height=15mm]{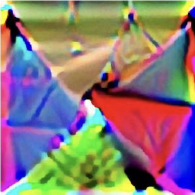}
    \end{minipage}
    \\ \hline %%%%%%%%%%%%%%%%%%%%%%%%%%%%%%%%%%%%%%%%%%%%%%%%%%%%%%%%%%%%%%%
    %Naturalness &  &  &  & \\
    Scores & 94.45\% & 47.75\% & 18.15\% & 15\%
        \\ \hline %%%%%%%%%%%%%%%%%%%%%%%%%%%%%%%%%%%%%%%%%%%%%%%%%%%%%%%%%%%%%%%
  \end{tabular}
\end{table}
Our proposed framework places a primary emphasis on the naturalness and inconspicuousness of the generated adversarial patch as perceived by human observers. To rigorously assess this aspect, we conducted a formal set of subjective evaluations, comparing the naturalness of our proposed patches against both baseline and real images. This subjective survey involved a diverse cohort of 20 participants, encompassing individuals of varied backgrounds, both male and female, aged between 19 and 59, and with no prior exposure to the topic. This comprehensive framework was taken to minimize potential biases and bolster the trustworthiness of our subjective evaluation.

Participants were tasked with providing a numerical score, ranging from 0 to 100\%, to gauge both the naturalness and the absence of conspicuous patterns in the presented images. As presented in Table \ref{tbl:nature}, our patch received the highest score (94.45\%) compared to NAP (47.75\%). This outcome underscores the patch's capacity to seamlessly integrate with its surroundings or context, as assessed by the participants.
\vspace{-4mm}
%============================
\section{Physical Attack Evaluations}
%============================

\begin{table}[!ht]
  %\small
    \centering
    \caption{Attack Success Rate (ASR) in Benign scenarios and when using AdvART and NAP \cite{Hu21} when attacking Yolov4tiny.}
    \begin{tabular}{cccc}
    \toprule
        \textbf{} & \textbf{Benign} & \textbf{NAP} & \textbf{AdvART} \\
        \midrule
         Attack Success Rate &  100\%   &  55\% & 72\%\\
    \bottomrule
    \end{tabular}    
    \label{tab:detection_tshirt}
\end{table}
We assessed the performance of our AdvART in real-world scenarios through the following setting: when the adversarial patch is printed on a t-shirt. By conducting evaluations in this physical context, we aimed to comprehensively evaluate the effectiveness of our technique.
%We print the patch and we test its performance in a real world setting.
We run a physical attack evaluation by taking pictures of one person wearing the adversarial t-shirt (see Figure \ref{advarttshirt}). We used YOLOv4tiny as the evaluation detector. Table \ref{tab:detection_tshirt} reports the detection recall in the physical world and when our AdvART is printed on a t-shirt, only 28\% of the time a person is detected.

\section{Discussion}
%=============================
%\subsection{Ablation Study}
%=============================
%Similarity Loss Ablation: Without the similarity loss function, the generated patch effectively becomes noise with conspicuous patterns. This highlights the pivotal role of the similarity loss in guiding the patch to resemble a benign image and enhancing its inconspicuousness. 

%Detection Loss Ablation: Without the detection loss, we find that the generated patch becomes remarkably similar to the target image. However, interestingly, this change does not lead to any discernible impact on the detector's performance. This suggests that the detection loss plays a crucial role in ensuring the patch's adversarial nature and its effectiveness in evading detection. 

%Total Variation Loss Ablation: Without the total variation loss function, we observe that the color change within the patch becomes sharper. This indicates that the total variation loss contributes significantly to the smoothness and natural appearance of the generated patch.

%=============================
\subsection{Influence of Different Similarity Metrics}
\begin{figure}[!ht]
\centering
\includegraphics[width=0.8\columnwidth]{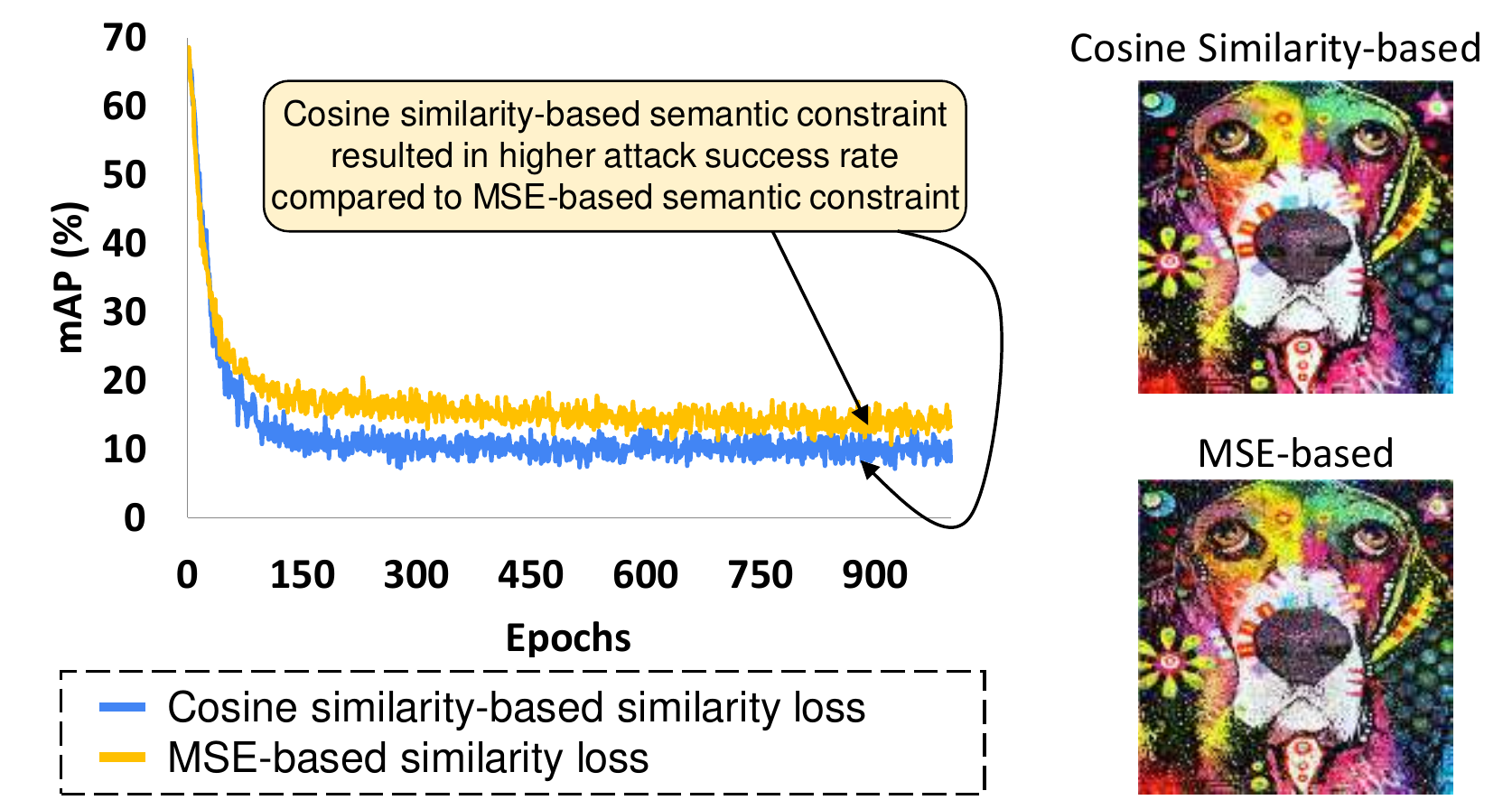}
\caption{Impact of different similarity metrics on the effectiveness of the generated patch.}
\label{different_sim_metrics}
\end{figure}

%=============================
We tested two different similarity metrics MSE and Cosine similarity used to measure the distance between two images to build our semantic constraint. As shown in Figure \ref{different_sim_metrics}, when using the cosine similarity the optimization function converged faster to a more efficient patch (i.e., lower mAP). The MSE-based achieved an mAP of 14\% however the cosine similarity based achieved a lower mAP of 8.81\%. 
In Table \ref{ssim}, the Structural Similarity Index (SSIM) \cite{wang2002universal} was computed between the benign target artistic pattern and the generated patch, utilizing both the Cosine similarity and MSE. This evaluation allowed us to assess the level of structural similarity between the original pattern and the generated adversarial patch, considering both their visual resemblance and semantic consistency. The final patch generated using the cosine similarity is more realistic and closer to the target benign pattern with an SSIM score of 0.95 compared to 0.88 for MSE based similarity loss.

 \begin{table}[!ht]
  \centering
%\small
  \caption{SSIM between the benign target artistic pattern and the generated patch using Cosine similarity and MSE-based semantic constraint.}
  \label{ssim}
  \begin{tabular}{ccc}
    \hline
    \textbf{Similarity metric} & \textbf{Cosine Similarity}  &\textbf{MSE} \\
    \hline
      \textbf{SSIM}  &   0.95   &   0.88     \\
  \hline
\end{tabular}
\end{table}

%\vspace{-4mm}
%=============================
\subsection{Influence of Different Transformations}
\begin{figure}[!ht]
\centering
\includegraphics[width=1\columnwidth]{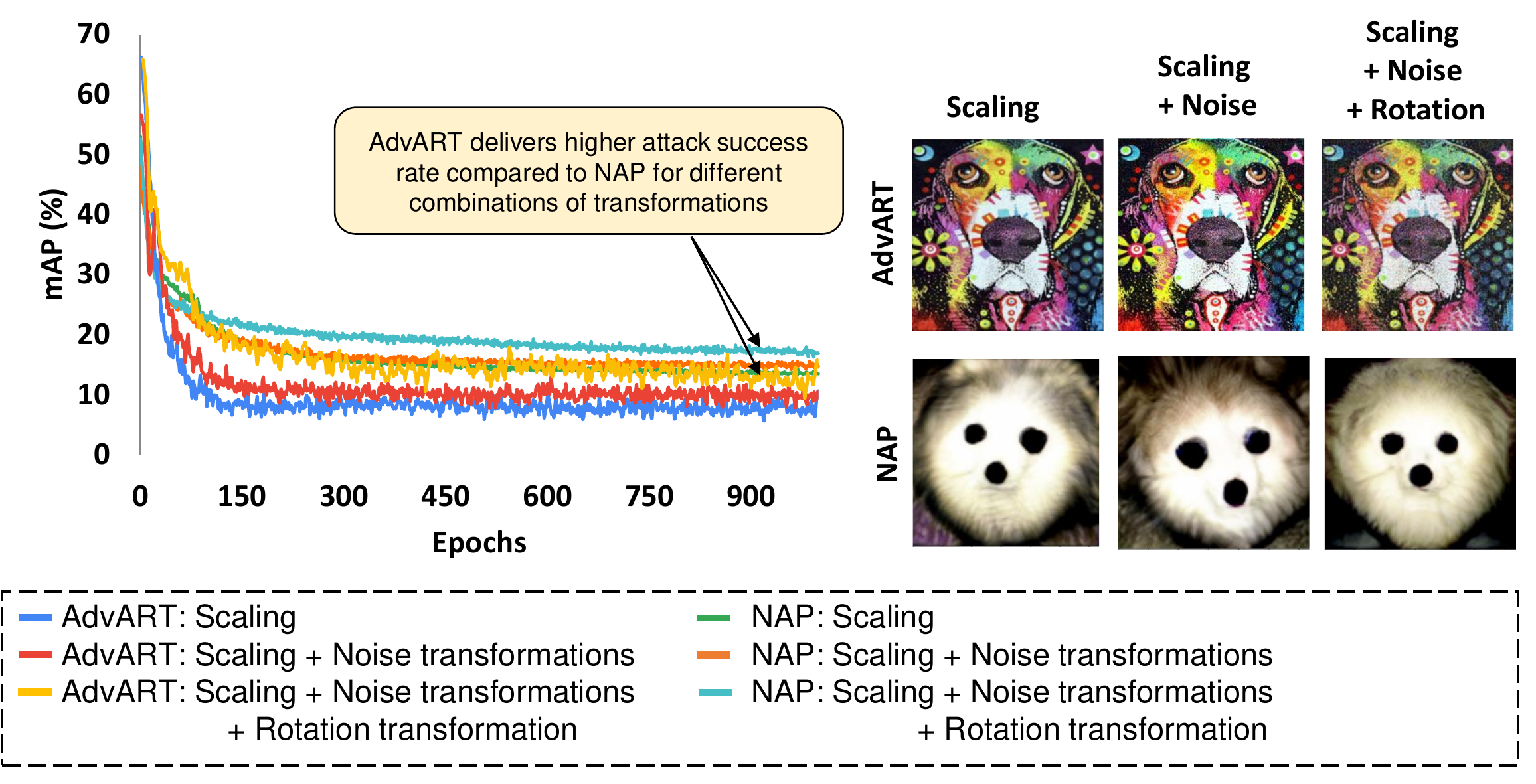}
\caption{Mean Average Precision (mAP) convergence curves when training a GAN-based technique (NAP) and our technique (AdvART) with different combinations of transformations.}
\label{different_transformation}
\end{figure}

%=============================
\label{sec:transformations}
%We explore various transformation methods for our scene rendering module to find suitable transformations that can enhance the attack robustness under different real-world distortions. 
To evaluate the capacity of our technique to withstand different set of transformations, we incorporate different combinations of transformations in both patch optimization process ours and the GAN-based; With random scaling, with random noise (Brightness, Contrast, and Gaussian Noise) and random scaling, and with random noise (Brightness, Contrast, and Gaussian Noise), random rotation, and random scaling. As shown in Figure \ref{different_transformation}, our technique outperforms the GAN-based technique and resulted in a higher attack success rate for different transformations, this could be explained by the too limited latent space of the GAN-based technique compared to our technique which provides higher flexibility since it's based on directly manipulating the pixels of the patch.
\vspace{-3mm}
\vspace{-4mm}
%=============================
\subsection{Influence of Patch Size}
%=============================
We run a number of digital experiments on the INRIA dataset to assess the performance influence of the patch size with respect to the size of the target object (person). Table \ref{patch_size} shows that the larger the size of the patch is, the stronger its attack performance as expected. 
 \begin{table}[!ht]
  \centering
%\small
  \caption{Attack performance of adversarial patches in different size settings for the INRIA dataset.}
  \label{patch_size}
  \begin{tabular}{lccc}
    \hline
    \textbf{Patch size} & \textbf{0.1}  &\textbf{0.2}  & \textbf{0.3}\\
    \hline
      \textbf{mAP}  &   55.7$\%$   &   8.81$\%$  &   5.32$\%$   \\
  \hline
\end{tabular}
\end{table}
%=============================
\subsection{AdvART based on Different Art Work}
%=============================
We also targeted other artistic patterns, we used the cosine similarity-based similarity loss to generate artistic adversarial patches. Our findings revealed that these patches retained a high performance, achieving an average success rate of 90\% on the INRIA dataset when using the Yolov4tiny object detector. For instance, when employing patch (a), the mean Average Precision (mAP) dropped significantly, reaching as low as 9.85\%.
 \begin{figure}[!htp]
\centering
\includegraphics[width=\columnwidth]{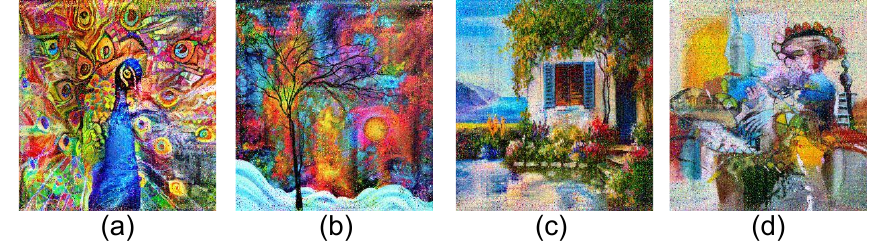}
\caption{Illustrations of AdvART based on different art work.}
\label{other_art_work}
\end{figure}
\vspace{-4mm}
 \begin{table}[!htp]
  \centering
%\small
  \caption{Attack performance of adversarial patches on different art work in Figure \ref{other_art_work} when attacking Yolov4tiny.}
  \label{other_art_work_map}
  \begin{tabular}{lcccc}
    \hline
    \textbf{Patch} & \textbf{(a)} &\textbf{(b)} &\textbf{(c)} &\textbf{(d)} \\
    \hline
      \textbf{mAP} &   9.85$\%$ &   10.2$\%$  &   11.33$\%$  &   15.14$\%$ \\
  \hline
\end{tabular}
\end{table}

\section{Conclusion}
\label{sec:conclusion}

In this paper, we introduce a novel framework for producing realistic/naturalistic and artistic physical adversarial patches for object detectors. AdvART is a framework that provides more flexibility in terms of integrated transformations when compared to GAN-based approaches, and it is capable of successfully producing realistic adversarial patches with natural/artistic patterns while retaining competitive attack performance compared to GAN-based techniques and non-naturalistic ones.
%==========================================================

\bibliographystyle{IEEEtran}
\bibliography{advnnbib}
%\balance
\end{document}